# Knowledge-Guided Machine Learning Models to Upscale Evapotranspiration in the U.S. Midwest

**Aleksei Rozanov**

rozan012@umn.edu

Department of Geography, Environment and Society,

GEMS Informatics Center,

Department of Computer Science and Engineering

University of Minnesota - Twin Cities

**Samikshya Subedi**

subed036@umn.edu

Department of Soil, Water, and Climate,

GEMS Informatics Center

University of Minnesota - Twin Cities

**Vasudha Sharma**

vasudha@umn.edu

Department of Soil, Water, and Climate

University of Minnesota - Twin Cities

**Bryan C. Runck***

runck014@umn.edu

GEMS Informatics Center

University of Minnesota - Twin Cities

*Corresponding Author

# Abstract

Evapotranspiration (ET) plays a critical role in the land-atmosphere interactions, yet its accurate quantification across various spatiotemporal scales remains a challenge. *In situ measurement approaches*, like eddy covariance (EC) or weather station-based ET estimation, allow for measuring ET at a single location. Agricultural uses of ET require estimates for each field over broad areas, making it infeasible to deploy sensing systems at each location. This study integrates tree-based and knowledge-guided machine learning (ML) techniques with multispectral remote sensing data, gridded meteorology and EC data to upscale ET across the Midwest United States. We compare four tree-based models - Random Forest, CatBoost, XGBoost, LightGBM - and a simple feed-forward artificial neural network in combination with features engineered using knowledge-guided ML principles. Models were trained and tested on EC towers located in the Midwest of the United States using k-fold cross validation with k=5 and site-year, biome stratified train-test split to avoid data leakage. Results show that LightGBM with knowledge-guided features outperformed other methods with an $R^2$=0.86, MSE=14.99 W·m$^{-2}$ and MAE = 8.82 W·m$^{-2}$ according to grouped k-fold validation (k=5). Feature importance analysis shows that knowledge-guided features were most important for predicting evapotranspiration. Using the best performing model, we provide a data product at 500 m spatial and one-day temporal resolution for gridded ET for the period of 2019-2024. Intercomparison between the new gridded product and state-level weather station-based ET estimates show best-in-class correspondence.

# Key Words

Evapotranspiration, agriculture, knowledge-guided machine learning, upscaling, MODIS, ERA5

# Highlights

1. We developed a tree-based machine learning model combining remote sensing, meteorology, and eddy covariance observations to upscale evapotranspiration (ET) at 500 m daily resolution, achieving high accuracy ($R^2$=0.86, MSE=14.99 W·m$^{-2}$ and MAE = 8.82 W·m$^{-2}$). Our model integrates knowledge-guided machine learning by incorporatingPenman-Monteith, demonstrating that physically based features significantly enhance ET prediction and dominate model feature importance rankings.

2. We generated a gridded ET dataset for the U.S. Midwest covering 2019–2024, offering consistent daily estimates suitable for agricultural applications, land surface modeling, and regional water management analysis.

# Introduction

Evapotranspiration (ET) represents the largest component of the crop water balance and serves as a key variable in irrigation planning and water resource management (Hargreaves, 1985; Pereira, 1999; Yang, 2023). Actual evapotranspiration(ETa) is amount of total crop-water loss from specific crop and field whereas reference evapotranspiration (ETr or ETo) is the hypothetical water loss of tall grass alfalfa (ETr) or short grass (ETo) from a well-watered reference surface with no water stress (Allen et al., 1994). Crop coefficient (Kc) plays a crucial role in irrigation water management, as they reflect varying water needs of crops at different crop-growth stages(Pereira, 2021). Efficient irrigation management that leverages accurate ETa estimates can significantly optimize water use, conserve resources, minimize environmental impacts, and promote sustainable agriculture (Ray, 2024; Subedi et al., 2025).This approach aligns with the broader goal of computationally designed agricultural research that connects data-driven models directly to farmer and stakeholder needs (Kantar et al., 2025). Aligning water supply with crop demand supports not only productivity but also water availability, quality, and long-term economic viability.

Despite the critical importance of ET, accurately estimating it remains a challenge due to limited ground-based observations. Their spatial coverage is inherently limited. In contrast, remote sensing has emerged as a powerful tool to scale ET estimation over large areas. Satellite platforms, particularly those in polar orbits, provide consistent spatial and temporal coverage that enables regional-to-global monitoring. Nevertheless, capturing the dynamic and continuous behavior of ET as a process involving water vapor movement across heterogeneous landscapes remains a significant technical and methodological challenge (Senay, 2022).

Different methods have been developed to estimate evapotranspiration (ET), ranging from empirical and physically based equations to data-driven models. : 1) Direct measurement -Traditional methods, such as lysimeters and eddy covariance (EC) towers, have advanced our understanding of ETa at field scales (Allen et al., 1991; Baldocchi et al., 2001). 2) Remote sensing approaches to estimate ETa such as OpenET which includes different models like ALEXI/DIsALEXI, eeMETRIC, geeSEBAL, SIMS and SEEBop (Anderson et al., 2007; Allen et al., 2007; Bastiaanssen et al., 1998; Pereira et al., 2020; Senay et al.,2013). 3) Weather based approaches like ETr/ ETo (Pemman) and Hargreaves and Samani (HS) model, which requires weather variables to calculate reference evapotranspiration (Hargreaves & Samani, 1985; Allen et al., 1994).The Penman-Monteith (PM) equation is widely regarded as one of the most accurate and theoretically robust approaches. Recommended by the FAO-56 guidelines, the PM equation combines radiative, aerodynamic, and thermodynamic parameters to estimate reference evapotranspiration (ETo). Despite its accuracy, the PM method requires multiple meteorological inputs, which may not always be available in data-scarce regions. To address this, simpler empirical models like the HS model require only a limited number of inputs (primarily temperature and extraterrestrial radiation) making it more suitable for areas with sparse observational networks. ETo is calculated from the nearest available weather station and then scaled to field-level ETa using Kc values for the target crop and growth stage. The HS equation (Eq. 1) is expressed as (Moeletsi, 2013):

$$ET_{HS} = 0.135 \cdot K_T \cdot (T_{avg} + 17.8) \cdot \sqrt{T_{max} - T_{min}} \cdot R_a \cdot 0.408 \cdot d, \qquad (1)$$

where $ET_{HS}$ is estimated ET (mm/dekad), $K_T$ – empirical coefficient , $T_{avg}, T_{max}, T_{min}$ are average, maximum, and minimum temperature (°C), $R_a$ – extraterrestrial radiation (MJ·m$^{-2}$), dd – number of days in the dekad and 0.408 is a conversion factor from energy to water depth.

While convenient, these empirical methods must be calibrated and validated against local climatic data and conditions to ensure reliability (Hargreaves & Samani, 1985; Moeletsi, 2013).Errors in weather-based methods arise both from the representativeness of the weather station (e.g., distance from the field) and from generalized Kc values, which may not capture local management or environmental variability.Eddy covariance (EC) method is standard methods for measuring how energy and water move between land and atmosphere. What sets EC apart is that it measures sensible (H) and latent (LE) heat fluxes directly rather than estimating from other weather variables. Despite their values, EC towers are scattered sparsely. Upscaling ETa measurements from EC towers to regional scales provides a more complete picture of energy and water exchanges, which helps us to understand how these systems respond to water and energy exchanges across the US Midwest and beyond (Jung et al., 2019; Xu et al.,2018).

## Machine Learning Approaches to Estimate ETa

In recent years, machine learning (ML) has emerged as a powerful alternative for estimating ETa, especially in data-limited environments. Unlike empirical models, ML algorithms do not rely on predefined functional forms. Instead, they learn complex and often nonlinear relationships from historical data, making them well-suited for modeling highly variable hydrometeorological processes.

ML models such as Multilayer Perceptrons (MLPs), Extreme Gradient Boosting (XGBoost), Light Gradient Boosting Machine (LightGBM), and CatBoost have demonstrated competitive or superior performance compared to traditional ET models across various climates (Liang et al., 2023; Prokhorenkova et al., 2018; Ke et al., 2017; Chen & Guestrin, 2016). A key advantage of ML methods is their ability to make accurate predictions even when some input variables are missing or incomplete, which is particularly valuable in operational settings with inconsistent meteorological records. For example, Malik et al. (2017) applied a Multilayer Perceptron to model monthly pan-evaporation in the Indian Central Himalayas. The MLPNN demonstrated strong generalization across two study sites and outperformed traditional climate-based models, particularly at higher elevations. This superior performance was attributed to the model's ability to capture nonlinear interactions among climatic variables, which are capabilities that are often limited in empirical approaches.

Another study by dos Santos, Robson Argolo, et al. (2024) evaluated the performance of 13 different ML algorithms for ET prediction across three distinct biomes, using remote sensing data from Landsat 8–9 and Sentinel-2 satellites. The study highlighted the sensitivity of model performance to both biome type and input data sources. It demonstrated that gradient-boosting models, especially CatBoost and LightGBM, performed consistently well across varying environmental conditions. The study also reinforced the value of multispectral imagery in characterizing vegetation and surface energy dynamics relevant to ET estimation.

While machine learning models have demonstrated remarkable performance in modeling complex environmental processes (including ETa), they often operate as black boxes, lacking physical interpretability and generalizability across regions or our out-of-sample observations. To address these limitations, recent research has explored Knowledge-Guided Machine Learning (KGML), which integrates domain knowledge from physical models with data-driven algorithms. There are three primary ways that domain knowledge is integrated into ML according to Karpatne et al., 2024: knowledge-guided learning, knowledge-guided architecture, and knowledge-guided pretraining. In knowledge-guided learning, domain knowledge is incorporated into the training algorithm (for example, by adding domain-based constraints or physics-inspired terms to the loss function) to steer the model towards scientifically consistent solutions. Knowledge-guided architecture means incorporating prior knowledge directly into the model's design – for instance, encoding known physical laws or invariances within the neural network structure itself. Finally, knowledge-guided pretraining leverages domain knowledge to initialize model parameters (often by pretraining on simulation data from scientific models before fine-tuning on real observations), thereby infusing scientific knowledge into the model. Summing up, KGML approaches combine the flexibility of machine learning with the scientific rigor of physically based equations to constrain or inform the learning process. This hybrid strategy enhances model interpretability, improves robustness in data-scarce regions, and leads to improved predictive accuracy (Karpatne et al., 2024; Runck et al., 2022).

This study aims to develop a high-resolution evapotranspiration (ET) modeling framework that integrates machine learning with knowledge-guided strategies using remote sensing, meteorological data, and eddy covariance (EC) observations. Our objectives are threefold: (1) to evaluate the performance of the KGML model using a rigorous grouped k-fold cross-validation approach across diverse ecosystem types in the U.S. Midwest; (2) to quantify the contribution of physical and remote sensing features to ET predictions through feature importance analysis; and (3) to generate a daily gridded ET product at 500 m resolution for the period of 2019–2024, offering a valuable dataset for agricultural management, hydrological modeling, and regional climate analysis.

# Materials and Methods

## Study Area

Our study focuses on the Central Midwest and Great Plains regions of the United States (Fig. 1), encompassing the states of Minnesota, Iowa, Wisconsin, North Dakota, South Dakota, Nebraska, Illinois, Indiana, Missouri, Michigan, and Kansas. This region forms the heart of U.S. agricultural production and has experienced a marked increase in irrigation use over the past two decades, as producers seek to expand cultivation into more marginal lands and mitigate risks associated with increasing drought frequency and climate variability. The climate across this area transitions from humid continental in the eastern and northern parts (e.g., Minnesota, Wisconsin, Michigan) to semi-arid in the Great Plains (e.g., Nebraska, Kansas, the Dakotas). These gradients contribute to substantial variability in evapotranspiration driven by differences in precipitation, temperature, wind speed, and solar radiation. The region experiences warm to hot summers, with peak ET rates occurring during the growing season from May through August. The study period spans from 2000 to 2024, covering a wide range of hydroclimatic conditions, including major drought events and increasing trends in summer heat extremes, making this dataset suitable for developing and validating robust, scalable ETa estimation models.

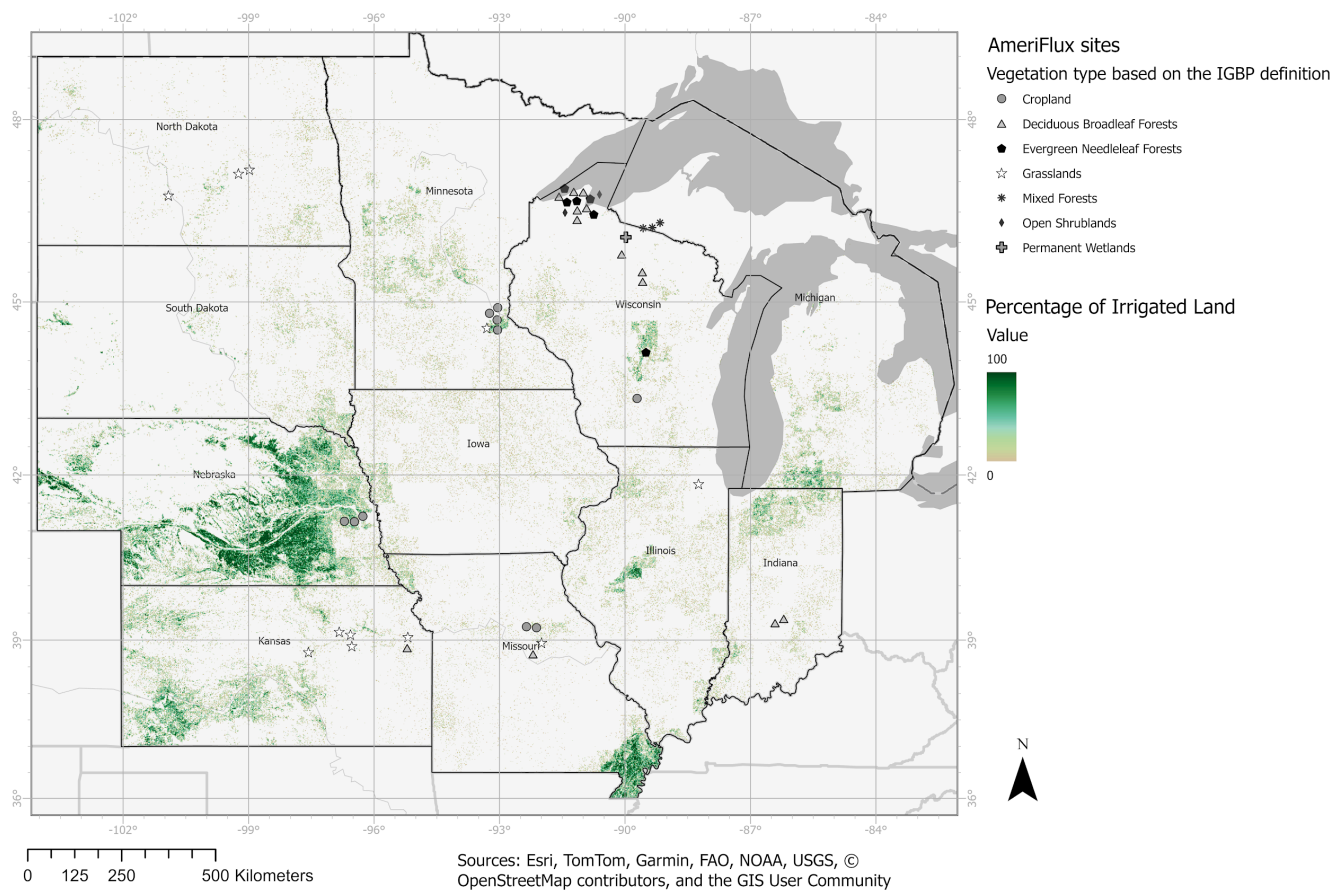

**Figure 1.** An overview of the Central Midwest and Great Plains Region USA, where the selected FLUXNET and AmeriFlux sites are located. To add more context to the study region, the percentage of irrigated land derived from MIrAD-US dataset by Pervez, Md Shahriar, and Jesslyn F. Brown. (2010) is also shown.

# Data

To solve the ET upscaling problem we developed a unique dataset consisting of four data sources: ERA5-Land (Muñoz Sabater, 2019), MOD09GA v061 (Vermote and Wolfe, 2021), AmeriFlux and FLUXNET (Pastorello et al., 2020). AmeriFlux and FLUXNET provide ground-based measurements of evapotranspiration and related fluxes using eddy covariance towers across a range of ecosystems.

ERA5-Land retrospective climate reanalysis is a global gridded dataset produced using 4dVAR data assimilation (Courtier, 1994) techniques providing information about terrestrial meteorology from 1950 onwards. In the current work, we used daily aggregates of ERA5-Land distributed by Google Earth Engine (Gorelic et al., 2017). The selected meteorology variables reflect the primary physical processes governing ET:

1. Air temperature and dewpoint temperature at 2 m control vapor pressure deficit, a key driver of atmospheric demand for ET.

2. Wind components and surface pressure influence turbulent transfer and enhance vapor removal at the surface, and affects air density, which modulates latent heat fluxes, respectively.

3. Surface net solar radiation directly supplies the energy needed for evaporation and transpiration.

4. Total evaporation and total precipitation represent a process-based model proxy for moisture fluxes, and water availability, setting boundary conditions for soil evaporation and plant transpiration respectively.

Since daily ET is not only defined by the current weather conditions, but also the meteorological events of the recent past (Kabala, 2025), all the meteorologies were derived as 30 days time series, where the last day in the array is the actual day of ET observation and other 29 characterise the weather in the recent past.

MOD09GA v061 provides estimates of surface spectral reflectance from the Moderate Resolution Imaging Spectroradiometer (MODIS) as it would appear at ground level without atmospheric scattering or absorption effects. The dataset is distributed as a daily gridded Level 2G product in a sinusoidal projection, offering 500 m reflectance data along with 1 km observation and geolocation statistics. Similar to ERA5-Land, MOD09GA v061 was accessed through Google Earth Engine (GEE). To extract proxies potentially related to ET, we included the following variables:

1. Sensor and viewing geometry (sensor zenith angle, sensor azimuth angle, solar zenith angle, solar azimuth angle), providing information about observation geometry and solar energy input.

2. Surface reflectance bands 1–7 (Red, NIR1, Blue, Green, NIR2, SWIR1, SWIR2), capturing vegetation and land surface conditions closely tied to ET processes.

3. Clouds QA (State 1km), which was binary decoded to flag pixels affected by clouds or shadows (1) versus clear observations (0).

Lastly, the target (dependent) variable for the supervised ML task was derived from AmeriFlux and FLUXNET eddy covariance (EC) time series at 38 sites (Fig. 1; Appendix A and B) across the U.S. Midwest, processed using the standardized OneFlux methodology (Pastorello et al., 2020). Specifically, we used LE_F_MDS (latent energy flux) as the dependent variable, pre-filtered to include only daily values with LE_F_MDS_QC $\geq$ 0.75, ensuring that observations with less than 25% gap-filling were retained.

Since evapotranspiration is directly related to latent energy (LE) through a simple energy-to-depth conversion (ET = LE / $\lambda$, where $\lambda$ is the latent heat of vaporization) (Pereira, 1998), LE serves as a physically consistent and robust proxy for ET in this study. Additionally, site latitude, longitude, IGBP (International Geosphere-Biosphere Programme; Loveland and Belward, 1997) ecosystem classification and observation day of year were extracted from the AmeriFlux and FLUXNET products and used as features in our own dataset.

# Feature Engineering

To enhance the predictive power of the future models, feature engineering is a step as important as feature selection. Different machine learning models vary in their ability to internally capture complex mathematical relationships. These relationships include, but are not limited to, counts, standard deviations, distances, and powers. If a model can learn such patterns inherently, manually crafting those features may be unnecessary, but according to (Heaton, 2016), many classical ML architectures struggle to capture certain relationships.

In the current work, we used three different approaches for feature engineering and extraction: First, we constructed weather-based time series features. Since ERA5-Land reanalysis was derived as a 30 days time series, this data contains temporal information that cannot be directly inferred by classical tree-based models. To extract the temporal information about meteorology, we computed minimum, maximum, and standard deviation values for each ERA5 variable in the dataset. For variables representing cumulative meteorological characteristics (e.g. surface net solar radiation, total evaporation, and total precipitation), we computed 7 and 30 days rolling sums, and for other variables, rolling means. These steps resulted in a time series dataset where each meteorological time series was transformed into a vector with the following components: last value, minimum, maximum, standard deviation, the rolling 30-day average or sum, and the rolling 7-day average or sum. Then all the vectors were stacked into one 1-dimensional vector.

Second, we constructed remote sensing-based features. Previous work (Running, 2004; Xiao, 2004; Rozanov and Gribanov, 2023) has shown that remote sensing vegetation indexes can be a strong predictor of terrestrial fluxes. In our current work, we used Normalized Difference Vegetation Index (NDVI; Eq. 2), Enhanced Vegetation Index (EVI; Eq. 3), Green Normalized Difference Vegetation Index (GNDVI; Eq. 4), Soil Adjusted Vegetation Index (SAVI; Eq. 5), Atmospherically Resistant Vegetation Index (ARVI; Eq. 6).

$$NDVI = \frac{NIR-Red}{NIR+Red}, \qquad (2)$$

$$EVI = \frac{NIR-Red}{NIR+6\,Red-7.5\,Blue+1}, \qquad (3)$$

$$GNDVI = \frac{NIR-Green}{NIR+Green}, \qquad (4)$$

$$SAVI = \frac{NIR-Red}{NIR+Red+0.5}1.5, \qquad (5)$$

$$ARVI = \frac{NIR-2Red+Blue}{NIR+2\,Red-BLue}, \qquad (6)$$

Lastly, constructed KGML features. In the context of ET upscaling, although the PM equation (Eq. 7) could be incorporated as a custom loss term, such an approach risks diverging from the observed measurements, given that PM does not fully capture the true variability of ET across different ecosystems (Choudhury, 1997). To mitigate this limitation of PM, while still leveraging prior physical knowledge, we incorporated PM-derived ET estimates computed from ERA5-Land meteorological features as input features to the machine learning models. Specifically, six new features were generated: the last recorded value, minimum, maximum, standard deviation, 30-day rolling mean, and 7-day rolling mean of PM-based ET time series.

$$ET_o = \frac{0.408\Delta(R_n - G) + \gamma \frac{900}{T+273} U_2 (e_s - e_a)}{\Delta + \gamma(1 + 0.34 U_2)}, \tag{7}$$

where $R_n$ is net radiation at the crop surface (MJ·m$^{-2}$·day$^{-1}$), G is soil heat flux density (MJ·m$^{-2}$·day$^{-1}$), T is air temperature at 2 m height (°C), U2 is wind speed at 2 m height (m·s$^{-1}$), $e_s$ is saturation vapor pressure (kPa), $e_a$ is Actual vapor pressure (kPa), $\Delta$ is slope of the vapor pressure curve (kPa·°C$^{-1}$), and $\gamma$ is psychrometric constant (kPa·°C$^{-1}$).

## Model Description

In the current work we attempted to use several recently developed tree-based ML architectures, such as Extreme Gradient Boosting (XGBoost; Chen et al., 2016), Categorical Boosting (CatBoost; Prokhorenkova et al., 2018) and LightGBM (Ke et al., 2017) against commonly-used baselines, Random Forest (RF; Breiman, 2001) and Artificial Neural Network (ANN), also known as Multilayer Perceptron (MLP; Rosenblatt, 1958). Fig. 2 gives a high-level overview of our methodology, including data acquisition, pre-processing and modeling itself. In the following, we describe each model and our implementation. Across all tree-based models, we selected optimal hyperparameters through grid search, unless otherwise noted.

First, RF is an ensemble of regression trees trained using a bootstrapping technique and was implemented with the scikit-learn (Pedregosa, 2011; ver. 1.6.1) Python package. Grid search resulted in 100 trees (n_estimators=100) and a maximum tree depth of 22 (max_depth=22).

Second, CatBoost is a gradient boosting algorithm designed to handle categorical features natively, employing ordered boosting to prevent target leakage and improve prediction stability. The model was implemented using the CatBoost (ver. 1.2.7) Python library with tree depth equal to 16, number of iterations set to 2000, and bootstrapping type equal to Poisson.

Third, XGBoost is a gradient boosting framework that builds additive tree models sequentially. It was trained with the following hyperparameters: a learning rate of 0.1, a maximum tree depth of 14, a minimum child weight of 5, and a gamma value of 0.6 to regularize tree splitting. Additionally, column subsampling (colsample_bytree=0.9) and row subsampling (subsample=0.8) were applied to enhance model generalization. The ensemble consisted of 100 trees (n_estimators=100). The model was trained using the xgboost (3.0.2) python package.

Four, LightGBM is a gradient boosting framework (ver. 4.6.0) based on decision tree algorithms, optimized for efficiency with histogram-based learning and leaf-wise tree growth. The model was tuned with 80 leaves, 0.05 learning rate, 0.7 colsample_bytree, optimizing RMSE.

Lastly, we built an MLP model with two hidden layers of 400 and 100 neurons and a tanh activation function applied to a weighted linear combination of each neuron using pytorch (Paszke, 2019; 2.6.0). Batch normalization and dropout with probability of 0.3 were used as well as regularization techniques to prevent overfitting.

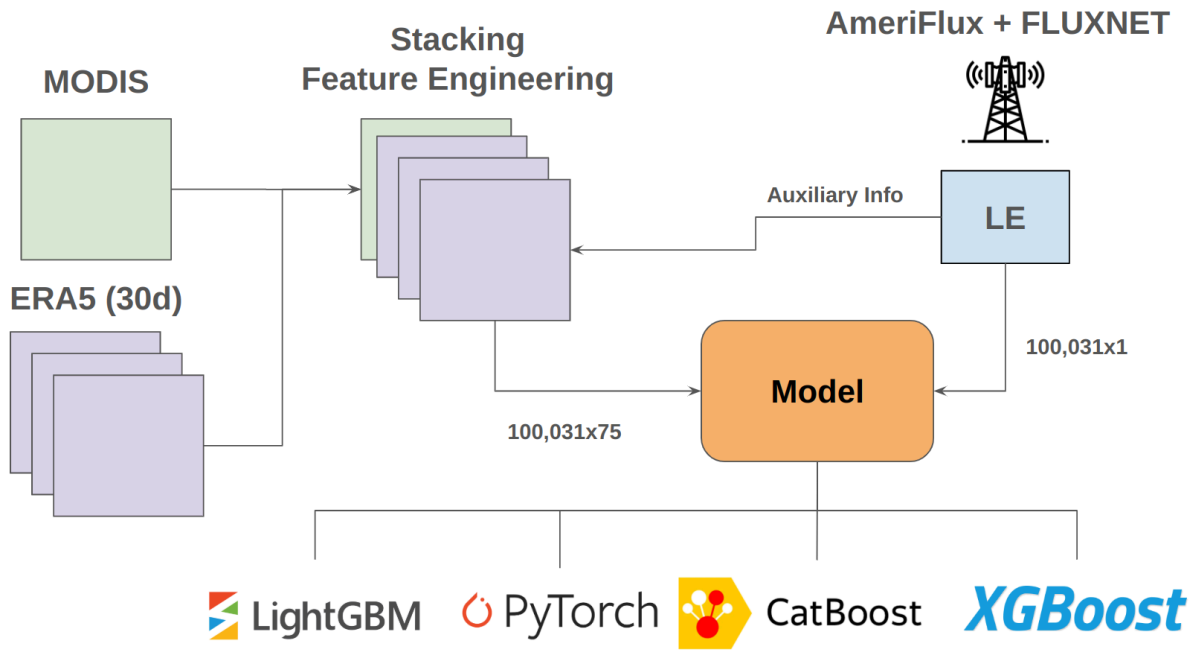

**Figure 2.** A data flow diagram, demonstrating the developed framework with data sources and processing stages applied to them.

# Results and Discussion

## Evaluation Metrics Selection

To compare models performance three common metrics were used: mean absolute error (**MAE**; Eq. 8), Root Mean Square Error (**RMSE**; Eq. 9), and **R²** (Eq. 10):

$$MAE = \frac{1}{n}\sum_{i=1}^{n} |y_{i,pred} - y_{i,true}|, \tag{8}$$

$$RMSE = \sqrt{\frac{1}{n}\sum_{i=1}^{n}(y_{i,pred} - y_{i,true})^2}, \tag{9}$$

$$R^2 = 1 - \frac{\sum_{i=1}^{n}(y_{i,pred} - \overline{y_{true}})^2}{\sum_{i=1}^{n}(y_{i,true} - \overline{y_{true}})^2}, \tag{10}$$

where $n$ is the number of data samples, $y_{pred}$ is a model's prediction and $y_{true}$ is the resembling ground-true value.

## Model Performance

To ensure rigorous model evaluation and prevent data leakage across spatial and temporal domains, we used GroupKFold (k=10) cross-validation from *scikit-learn* python library with site-year as a grouping variable. This approach ensures that all observations from a given site and year are confined to either the training or validation set, never both. Such partitioning prevents information from leaking across temporal or spatial boundaries, which is particularly important in flux upscaling tasks where environmental variables exhibit strong site-level and seasonal autocorrelations. Traditional train-test splits, even when stratified by biome (IGBP), risk overestimating performance by allowing similar conditions from the same site or time period to influence both training and testing. In contrast, our group-based cross-validation yields higher error than train-test split but provides a more realistic estimate of generalization performance, simulating the model's ability to extrapolate to entirely unseen site-year conditions. We discarded all other validation strategies that did not explicitly account for grouping, as they can produce biased metrics and inflate confidence in model predictions.

Model performance is summarized in Table 1. LightGBM outperformed all other models achieving the lowest RMSE = 14.99 W·m$^{-2}$ (0.861 mm day$^{-1}$) and MAE = 8.82 W·m$^{-2}$ (0.528 mm day$^{-1}$), and the highest $R^2$ = 0.86 during the cross validation. CatBoost and XGBoost demonstrated similar performance, yet worse than LightGBM, whereas RF and ANN had the weakest performance on the given task.

**Table 1**. Comparison of the trained models on the test dataset from the site-year split.

| Model | Metric | | |
|---|---|---|---|
| | $R^2$ | RMSE, W·m$^{-2}$ | MAE, W·m$^{-2}$ |
| Random Forest | 0.838 | 16.07 | 9.585 |
| CatBoost | 0.832 | 16.374 | 9.629 |
| XGBoost | 0.840 | 15.985 | 9.46 |
| LightGBM | **0.860** | **15.990** | **8.823** |
| ANN | 0.756 | 19.720 | 13.444 |

In addition to global metrics, we evaluated model performance across individual months and IGBP land cover classes (Fig. 3) to assess its temporal and ecological generalization capabilities. Monthly error analysis showed that May, June, and July, corresponding to the period of highest ET flux, exhibited the largest errors, with RMSE values of 22.07 ± 0.49 W·m$^{-2}$ (0.775 ± 0.017 mm day$^{-1}$), 25.85 ± 1.09 W·m$^{-2}$ (0.913 ± 0.039 mm day$^{-1}$), and 23.58 ± 0.36 W·m$^{-2}$ (0.835 ± 0.012 mm day$^{-1}$) respectively (± indicates the standard error across folds). In contrast, winter months (December, January, and February) showed substantially lower errors and variability, with RMSEs of 3.86 ± 0.265 W·m$^{-2}$ (0.133 ± 0.009 mm day$^{-1}$), 3.8 ± 0.053 W·m$^{-2}$ (0.132 ± 0.002 mm day$^{-1}$), and 5.06 ± 0.14 W·m$^{-2}$ (0.175 ± 0.005 mm day$^{-1}$), respectively.

When stratified by IGBP land cover class, the model predicted ET in Mixed Forests (MF) most accurately, with an RMSE of 6.462 ± 0.69 W·m$^{-2}$ (0.227 ± 0.024 mm day$^{-1}$). Higher errors were observed in Croplands (CRO), Wetlands (WET), and Grasslands (GRA), which had RMSE values of 16.09 ± 0.75 W·m$^{-2}$ (0.556 ± 0.027 mm day$^{-1}$), 17.718 ± 0.33 W·m$^{-2}$ (0.622 ± 0.012 mm day$^{-1}$), and 16.34 ± 0.374 W·m$^{-2}$ (0.576 ± 0.013 mm day$^{-1}$), respectively. These elevated errors likely reflect greater heterogeneity in canopy structure and soil moisture dynamics within these ecosystems.

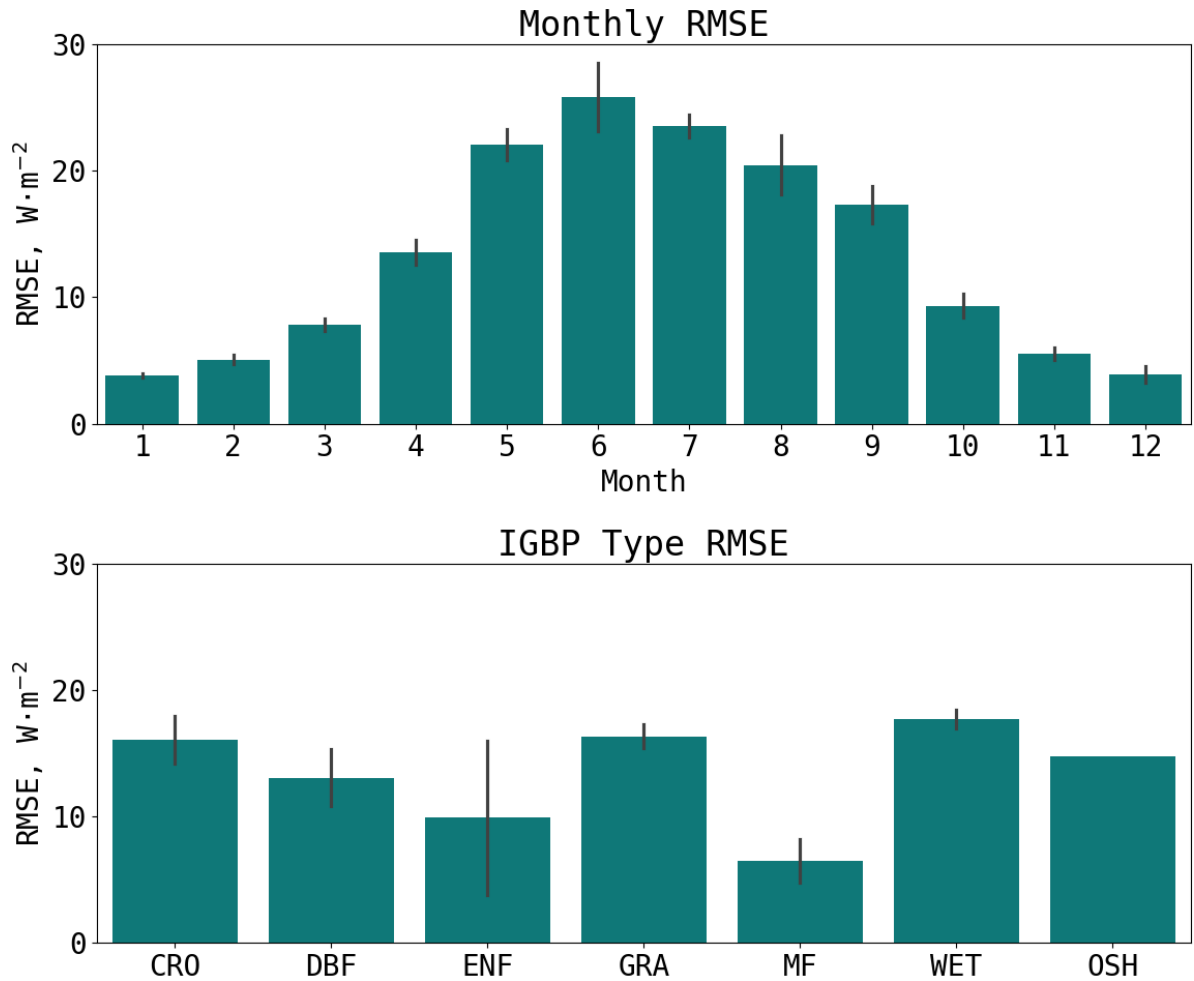

**Figure 3.** Error by month and IGBP type. CRO, DBF, ENF, GRA, MF, WET and OSH represent Croplands, Deciduous Broadleaf Forests, Evergreen Needleaf Forests, Grasslands, Mixed Forests, Wetlands and Open Shrublands, respectively.

Lastly, we conducted a feature importance analysis based on gain using the native LightGBM tools (Fig. 4). Feature importance by gain quantifies the contribution of each feature to reducing the overall loss, specifically RMSE, by aggregating the improvements made at each split where the feature is used across all trees. The results align with established understanding of the primary drivers of LE (and ET), highlighting the critical role of historical radiation, air temperature, and water balance in shaping ET dynamics across sites. Notably, the most important feature was the knowledge-guided input derived from the Penman-Monteith equation, demonstrating that the model strongly leverages physically based prior knowledge. This finding directly supports one of the core objectives of this study: to evaluate whether incorporating physical process knowledge enhances model performance. The results demonstrate that the most accurate model consistently prioritized the physically informed feature when making ET predictions.

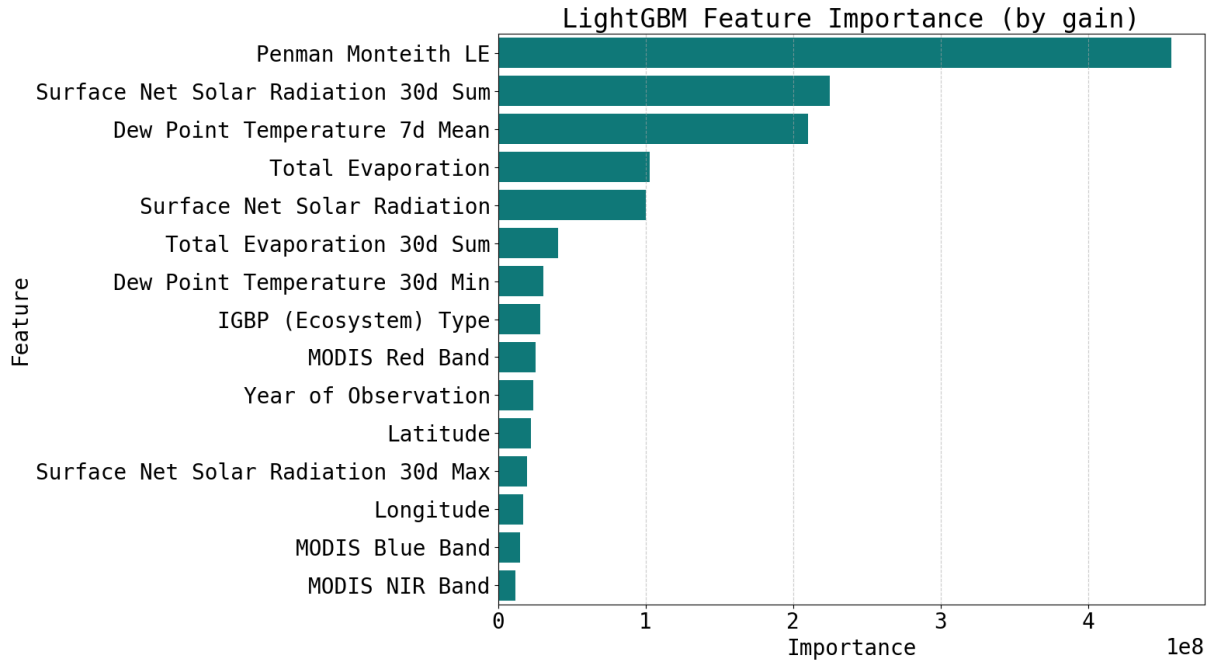

**Figure 4**. Feature importance analysis performed for the LightGBM model, where importances are computed based on the information gain at each split which is based on the given feature. The top 15 most impactful features are shown.

# Data Product

To support the needs of decision-makers in the U.S. Upper Midwest, we release an open-access dataset (Fig. 5) generated by our framework for the period 2019–2024. The data have a spatiotemporal resolution of 500 m and 1 day, covering latitudes 36°N to 49°N and longitudes −104°W to −82°W. The dataset is distributed in NetCDF4 format.

To ensure the quality of the generated data, we performed an additional evaluation by comparing monthly aggregated values with (i) the OpenET ensemble median product (Melton et al., 2021) obtained via Google Earth Engine (OpenET/ENSEMBLE/CONUS/GRIDMET/MONTHLY/v2_0), and (ii) Penman–Monteith ETr estimates derived from ground-based meteorological observations at 13 Minnesota Mesonet sites. The results (Fig. 6) show strong agreement with OpenET (r = 0.94, RMSE = 15.24 mm month$^{-1}$), reasonable agreement with the PM-based approach (r = 0.89, RMSE = 35.52 mm month$^{-1}$), and tighter correspondence with the PM-based approach than OpenET (r = 0.88; RMSE = 34.9 mm month$^{-1}$) a finding that is useful for local decision-makers evaluating data products for decision-support.

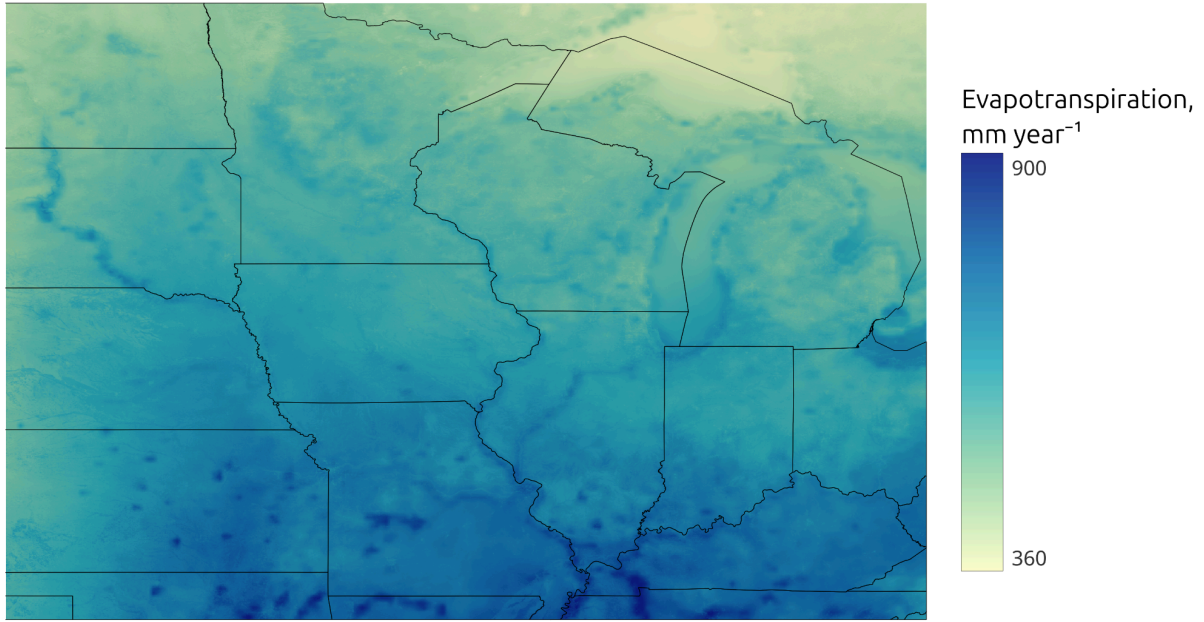

**Figure 5**. Average annual ET between 2019 and 2024 derived from the upscaled data product.

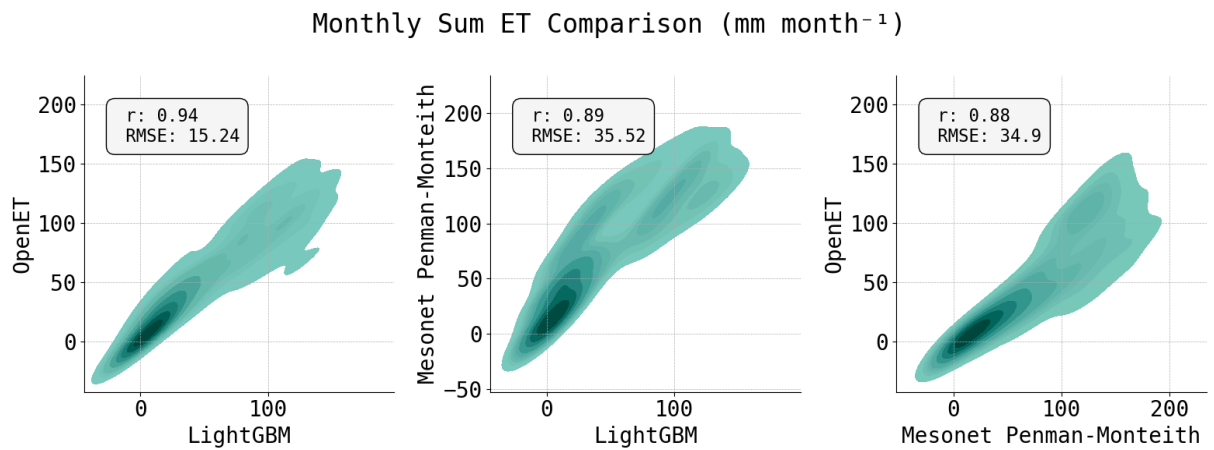

**Figure 6**. Intercomparison of monthly sum ET (mm month$^{-1}$) with ensemble median from OpenET and PM-derived ET based on the Mesonet stations meteorologies for 13 sites in Minnesota, US.

# Conclusion

Here, we presented a high-resolution (500 m, daily) ET product for the U.S. Upper Midwest, generated using a LightGBM-based machine learning model informed by physical principles. The model was rigorously validated using GroupKFold cross-validation by site-year, ensuring realistic assessments of generalization performance and minimizing spatial and temporal leakage—an often-overlooked issue in environmental upscaling. Among evaluated methods, LightGBM demonstrated superior predictive accuracy across sites, seasons, and land cover types, with the highest $R^2$ (0.855) and lowest RMSE (15.2 $W·m^{-2}$). Model performance declined during peak ET months and over heterogeneous land cover types, consistent with increased flux variability.

Our results underscore the value of hybrid approaches: the model's top feature was a knowledge-guided input derived from the Penman–Monteith equation, highlighting the synergistic benefit of combining process understanding with data-driven methods. This demonstrates that physically informed features can meaningfully improve performance and interpretability in machine learning-based flux modeling.

To support practical applications, we publicly release the full ET dataset for 2019–2024 as an open-access resource: https://doi.org/10.13020/37XE-QQ18. The product shows strong agreement with independent benchmarks, including OpenET and ground-based PM estimates, confirming its utility for operational and research use. The dataset can inform irrigation planning, drought monitoring, and hydrological modeling across the Midwest.

Additionally, to promote transparency, open-source science, and research reproducibility, we make available all Python code developed during this study in a public Git repository: https://github.com/RTGS-Lab/ET_LCCMR. This includes data preprocessing workflows, model training scripts, and evaluation routines, enabling other researchers to build on our work and adapt the framework to new regions, fluxes, or remote sensing inputs.

# CRediT Authorship Contribution Statement

**Aleksei Rozanov:** Conceptualization, Methodology, Formal analysis, Data curation, Writing – original draft, review & editing, Visualization. **Samikshya Subedi:** Writing – original draft, review & editing. **Vasudha Sharma:** Writing – review& editing. **Bryan Runck:** Writing – review & editing, Conceptualization, Resources, Methodology, Project administration.

# AI Usage Statement

Portions of this manuscript, including language refinement and clarity improvements, were assisted by OpenAI's ChatGPT. All content was reviewed and edited by the authors to ensure accuracy and originality.

# Declaration of competing interests

The authors declare that they have no known competing financial interests or personal relationships that could have appeared to influence the work reported in this paper.

# Acknowledgements

Funding for this project was provided by the Minnesota Environment and Natural Resources

Trust Fund as recommended by the Legislative-Citizen Commission on Minnesota Resources

(LCCMR) ENRTF 2021-266.

# Data Availability

The dataset is openly available at the Data Repository for the University of Minnesota (DRUM): https://conservancy.umn.edu/items/ada05f34-e8a6-492f-83dd-04db7983ac19. The code created during this study is available through GitHub: https://github.com/RTGS-Lab/ET_LCCMR.

# Supplemental Materials

**Appendix A.** FLUXNET sites used in this study.

| Site | Link | Latitude | Longitude |
| --- | --- | --- | --- |
| US-IB2 | https://fluxnet.org/sites/siteinfo/US-IB2 | 41.8406 | -88.241 |
| US-Los | https://fluxnet.org/sites/siteinfo/US-Los | 46.0827 | -89.9792 |
| US-MMS | https://fluxnet.org/sites/siteinfo/US-MMS | 39.3232 | -86.4131 |
| US-Ne1 | https://fluxnet.org/sites/siteinfo/US-Ne1 | 41.1651 | -96.4766 |
| US-Ne2 | https://fluxnet.org/sites/siteinfo/US-Ne2 | 41.1649 | -96.4701 |
| US-Ne3 | https://fluxnet.org/sites/siteinfo/US-Ne3 | 41.1797 | -96.4397 |
| US-Syv | https://fluxnet.org/sites/siteinfo/US-Syv | 46.242 | -89.3477 |
| US-WCr | https://fluxnet.org/sites/siteinfo/US-WCr | 45.8059 | -90.0799 |
| US-Wi0 | https://fluxnet.org/sites/siteinfo/US-Wi0 | 46.6188 | -91.0814 |
| US-Wi1 | https://fluxnet.org/sites/siteinfo/US-Wi1 | 46.7305 | -91.2329 |
| US-Wi2 | https://fluxnet.org/sites/siteinfo/US-Wi2 | 46.6869 | -91.1528 |
| US-Wi3 | https://fluxnet.org/sites/siteinfo/US-Wi3 | 46.6347 | -91.0987 |
| US-Wi4 | https://fluxnet.org/sites/siteinfo/US-Wi4 | 46.7393 | -91.1663 |
| US-Wi6 | https://fluxnet.org/sites/siteinfo/US-Wi6 | 46.6249 | -91.2982 |
| US-Wi8 | https://fluxnet.org/sites/siteinfo/US-Wi8 | 46.7223 | -91.2524 |

**Appendix B.** AmeriFlux sites used in this study.

| Site | Link | Latitude | Longitude |
| --- | --- | --- | --- |
| US-CS2 | https://ameriflux.lbl.gov/sites/siteinfo/US-CS2 | 44.1467 | -89.5002 |

| Site | URL | Lat | Lon |
|---|---|---|---|
| US-DFC | https://ameriflux.lbl.gov/sites/siteinfo/US-DFC | 43.3448 | -89.7117 |
| US-KFS | https://ameriflux.lbl.gov/sites/siteinfo/US-KFS | 39.0561 | -95.1907 |
| US-KLS | https://ameriflux.lbl.gov/sites/siteinfo/US-KLS | 38.7745 | -97.5684 |
| US-Kon | https://ameriflux.lbl.gov/sites/siteinfo/US-Kon | 39.0824 | -96.5603 |
| US-MMS | https://ameriflux.lbl.gov/sites/siteinfo/US-MMS | 39.3232 | -86.4131 |
| US-MO1 | https://ameriflux.lbl.gov/sites/siteinfo/US-MO1 | 39.2298 | -92.1167 |
| US-MO2 | https://ameriflux.lbl.gov/sites/siteinfo/US-MO2 | 38.9488 | -91.9945 |
| US-MO3 | https://ameriflux.lbl.gov/sites/siteinfo/US-MO3 | 39.2322 | -92.1493 |
| US-MOz | https://ameriflux.lbl.gov/sites/siteinfo/US-MOz | 38.7441 | -92.2000 |
| US-Ro1 | https://ameriflux.lbl.gov/sites/siteinfo/US-Ro1 | 44.7143 | -93.0898 |
| US-Ro2 | https://ameriflux.lbl.gov/sites/siteinfo/US-Ro2 | 44.7288 | -93.0888 |
| US-Ro4 | https://ameriflux.lbl.gov/sites/siteinfo/US-Ro4 | 44.6781 | -93.0723 |
| US-Ro5 | https://ameriflux.lbl.gov/sites/siteinfo/US-Ro5 | 44.691 | -93.0576 |
| US-Ro6 | https://ameriflux.lbl.gov/sites/siteinfo/US-Ro6 | 44.6946 | -93.0578 |
| US-Syv | https://ameriflux.lbl.gov/sites/siteinfo/US-Syv | 46.2420 | -89.3477 |
| US-Wi0 | https://ameriflux.lbl.gov/sites/siteinfo/US-Wi0 | 46.6188 | -91.0814 |
| US-Wi1 | https://ameriflux.lbl.gov/sites/siteinfo/US-Wi1 | 46.7305 | -91.2329 |
| US-Wi3 | https://ameriflux.lbl.gov/sites/siteinfo/US-Wi3 | 46.6347 | -91.0987 |
| US-Wi4 | https://ameriflux.lbl.gov/sites/siteinfo/US-Wi4 | 46.7393 | -91.1663 |
| US-Wi6 | https://ameriflux.lbl.gov/sites/siteinfo/US-Wi6 | 46.6249 | -91.2982 |
| US-Wi8 | https://ameriflux.lbl.gov/sites/siteinfo/US-Wi8 | 46.7223 | -91.2524 |

| Site | URL | Latitude | Longitude |
|---|---|---|---|
| US-xDC | https://ameriflux.lbl.gov/sites/siteinfo/US-xDC | 47.1617 | -99.1066 |
| US-xKA | https://ameriflux.lbl.gov/sites/siteinfo/US-xKA | 39.1104 | -96.6129 |
| US-xKZ | https://ameriflux.lbl.gov/sites/siteinfo/US-xKZ | 39.1008 | -96.5631 |
| US-xNG | https://ameriflux.lbl.gov/sites/siteinfo/US-xNG | 46.7697 | -100.9154 |
| US-xST | https://ameriflux.lbl.gov/sites/siteinfo/US-xST | 45.5089 | -89.5864 |
| US-xTR | https://ameriflux.lbl.gov/sites/siteinfo/US-xTR | 45.4937 | -89.5857 |
| US-xUK | https://ameriflux.lbl.gov/sites/siteinfo/US-xUK | 39.0404 | -95.1921 |
| US-xUN | https://ameriflux.lbl.gov/sites/siteinfo/US-xUN | 46.2339 | -89.5373 |
| US-xWD | https://ameriflux.lbl.gov/sites/siteinfo/US-xWD | 47.1282 | -99.2414 |